\documentclass[journal,10pt,twocolumn,twoside]{IEEEtran}
\usepackage{times,amsmath,amssymb}
\usepackage{slashbox}
\usepackage{graphicx}
\usepackage{xcolor}
\usepackage{algorithm}
\usepackage{algorithmicx}
\usepackage{setspace}
\usepackage[export]{adjustbox}
\usepackage{multirow}
\usepackage[noend]{algpseudocode}
\usepackage{tabularx,booktabs}
\newcolumntype{Y}{>{\centering\arraybackslash}X}
\usepackage{tikz}
\newcommand{\Cross}{\mathbin{\tikz [x=1.4ex,y=1.4ex,line width=.2ex] \draw (0,0) -- (1,1) (0,1) -- (1,0);}}%
\newcommand*{\QEDA}{\hfill\ensuremath{\blacksquare}}%
\newcommand{\myvec}[1]{\ensuremath{\begin{bmatrix}#1\end{bmatrix}}}

\begin{document}

\title{Heterogeneous Multilayer Generalized Operational Perceptron}

\author{\IEEEauthorblockN{Dat Thanh Tran\IEEEauthorrefmark{1}, Serkan Kiranyaz\IEEEauthorrefmark{2}, Moncef Gabbouj\IEEEauthorrefmark{1}, and Alexandros Iosifidis\IEEEauthorrefmark{3},}\\
\IEEEauthorblockA{\IEEEauthorrefmark{1}Department of Computing Sciences, Tampere University, Tampere, Finland\\
\IEEEauthorrefmark{2}Department of Electrical Engineering, Qatar University, Qatar\\
\IEEEauthorrefmark{3}Department of Engineering, Electrical \& Computer Engineering, Aarhus University, Aarhus, Denmark\\
Email:{thanh.tran, moncef.gabbouj}@tuni.fi, mkiranyaz@qu.edu.qa, alexandros.iosifidis@eng.au.dk}\\
}

\maketitle

\begin{abstract}
The traditional Multilayer Perceptron (MLP) using McCulloch-Pitts neuron model is inherently limited to a set of neuronal activities, i.e., linear weighted sum followed by nonlinear thresholding step. Previously, Generalized Operational Perceptron (GOP) was proposed to extend conventional perceptron model by defining a diverse set of neuronal activities to imitate a generalized model of biological neurons. Together with GOP, Progressive Operational Perceptron (POP) algorithm was proposed to optimize a pre-defined template of multiple homogeneous layers in a layerwise manner. In this paper, we propose an efficient algorithm to learn a compact, fully heterogeneous multilayer network that allows each individual neuron, regardless of the layer, to have distinct characteristics. Based on the complexity of the problem, the proposed algorithm operates in a progressive manner on a neuronal level, searching for a compact topology, not only in terms of depth but also width, i.e., the number of neurons in each layer. The proposed algorithm is shown to outperform other related learning methods in extensive experiments on several classification problems.
\end{abstract}

\begin{IEEEkeywords}
	Generalized Operational Perceptron,
	Feedforward Network,
	Architecture Learning,
	Progressive Learning
\end{IEEEkeywords}

\section{Introduction}\label{S:Intro}
In recent years, learning systems based on neural networks have gained tremendous popularity in a variety of application domains such as machine vision, natural language processing, biomedical analysis or financial data analysis \cite{redmon2016you, iosifidis2012view, tsantekidis2017forecasting, graves2013speech, girshick2014rich, zabihi2016heart, tsantekidis2017using, hinton2012deep, waris2017cnn, an2014deep}. The recent resurgence of neural networks, especially deep neural networks, can be attributed to the developments of specialized computing hardware such as Graphical Processing Units (GPU) and improved training techniques or architecture designs such as Batch Normalization \cite{ioffe2015batch}, Dropout \cite{srivastava2014dropout}, Residual Connection \cite{he2016deep} as well as stochastic optimization algorithms such as Nesterov SGD \cite{bengio2013advances} or Adam \cite{kingma2014adam}, to name a few. In order to improve the performance and generalization capacity of neural networks, a large amount of effort has been made to learn deeper and deeper network topologies with larger, heavily annotated datasets. While the network architectures and training heuristics have evolved over the past few years, the core components of a neural network, i.e., the neuron model has remained relatively unchanged. The most typical artificial neuron model is based on McCulloch-Pitts perceptron \cite{mcculloch1943logical}, hereafter simply referred to as perceptron, which performs a linear summation with learnable synaptic weights followed by an element-wise nonlinear activation function. This design principle was aimed to \textit{loosely} simulate biological neurons in mammalian nervous system, and is used in the current state-of-the-art architectures such as Convolutional Neural Network (CNN) or Recurrent Neural Network (RNN).

\begin{figure}[b!]
	\centering
	\includegraphics[width=0.45\textwidth]{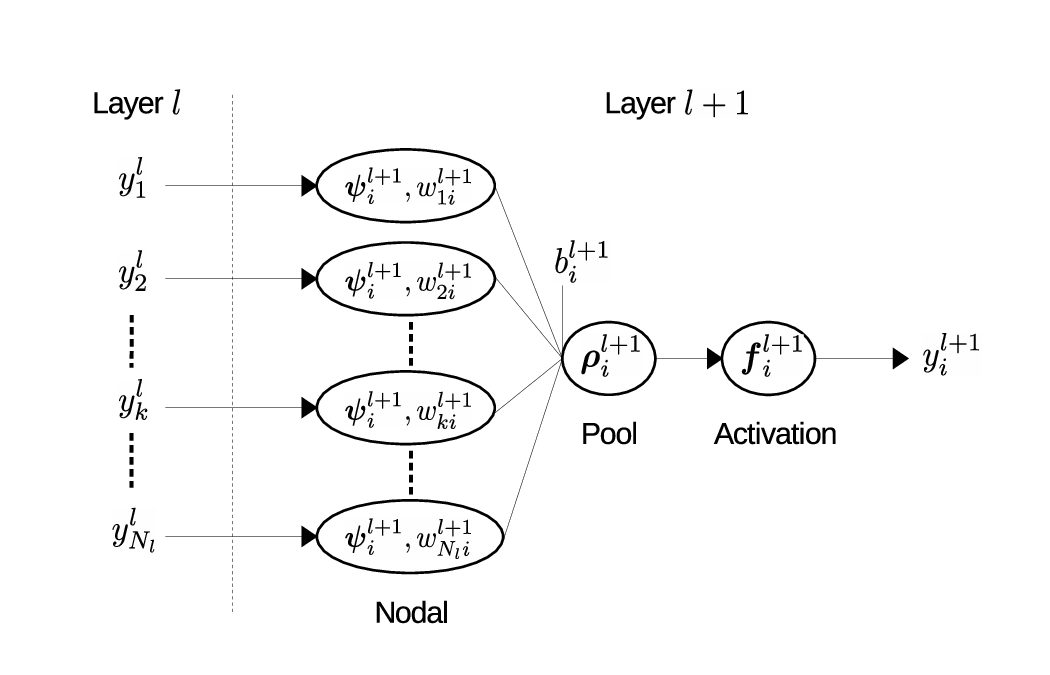}
	\caption{Computation of the $i$-th GOP neuron at layer $l+1$, characterized by the synaptic weights $w^{l+1}_{ki}$, the nodal operator $\boldsymbol{\psi}^{l+1}_i$, the pooling operator $\boldsymbol{\rho}^{k+1}_{i}$ and the activation operator $\boldsymbol{f}^{l+1}_i$}
	\label{f1}
\end{figure}

It was recently proposed in \cite{kiranyaz2017progressive} that the crude model of biological neuron based on McCulloch-Pitts design should be replaced with a more general neuron model called Generalized Operational Perceptron (GOP), which also includes the conventional perceptron as a special case. Each GOP is characterized by learnable synaptic weights and an operator set comprising of three types of operations: nodal operation, pooling operation, and activation operation, as illustrated in Figure \ref{f1}. The form of each operation is selected from a library of pre-defined operations. By allowing different neurons to have different nodal, pool and activation operators, GOP can encapsulate a diversity of both linear and nonlinear operations.

The diversity introduced by GOP poses a much more challenging problem as compared to the traditional perceptron: optimizing both the synaptic weights and the choice of the operator set. In order to build a neural network using GOP, a progressive learning algorithm called Progressive Operational Perceptron (POP) was proposed in \cite{kiranyaz2017progressive}, which optimizes a pre-defined network template in a layer-wise manner. To avoid an intractable search process in the combinatorial space of operator sets, POP constrains all GOPs within the same layer to share the same operator set. While limiting the functional form of neurons in the same layer, training POP is still painstakingly slow since the evaluation of an operator set involves training a single hidden layer network and the combinatorial search space of the optimal operator set in the output layer in conjunction with the hidden layer is enormous. POP achieves a partial degree of self-organization by adding new hidden layers only until the objective is met on the data forming the training set. The width of each layer is, however, pre-defined beforehand, leading to a suboptimal final topology in terms of compactness.

When building a learning system based on neural networks, the architectural choice of the networks' topology plays an important role in the generalization capacity \cite{ellis1999size}. The size of the neural network, i.e., the number of layers and the number of neurons in each layer, are usually selected based on some standard widely used structures or by manual tuning through experimentation. While some designs favor network depth such as Residual Networks with hundred of layers, empirical experiments in \cite{zagoruyko2016wide} have shown that shallower but wider topologies can achieve better generalization performance and computation efficiency. In case of densely connected topologies, an MLP with large layers can easily lead to overfitting while randomized neural networks \cite{pao1992functional, schmidt1992feedforward, huang2006extreme, rahimi2008random} typically formed by a large hidden layer with random neurons are robust to overfitting. There have been several attempts \cite{chatterjee2017progressive, chen2018broad, kiranyaz2017progressive, ivakhnenko1971polynomial, bengio2007greedy, kulkarni2017layer, stanley2002evolving, kiranyaz2009evolutionary, zoph2016neural} to systematically search for the optimized network architectures with a given objective criterion. Regarding densely connected networks, existing literature employs different learning strategies such as progressively adding neurons and solving for the synaptic weights by randomization or convex optimization, or both. While randomization and convex optimization are characterized with fast training time and usually come with some forms of theoretical guarantee, the resulting architectures are often large with hundreds or thousands neurons. Another popular approach is the application of evolutionary strategies in the architectural search procedure. For example, by encoding the network configurations and parameters into particles, multidimensional particle swam optimization was used to evolve both the network configurations and synaptic weights in \cite{kiranyaz2009evolutionary}. While evolutionary algorithms work well in practice, their fitness evaluation step often requires heavy computation, rendering their application in large datasets intractable. Recently, by employing powerful commodity hardware with $800$ GPUs, it was possible to evolve LSTM architectures on the Penn Treebank language modeling dataset \cite{zoph2016neural}. The common drawback of all the aforementioned learning systems is the use of perceptron model, which limits the learning capability of each neuron.

Due to the availability of low-cost embedded, mobile devices that are affordable to many customer classes, more and more research efforts have been focused on efficient inference systems on mobile devices that require small memory footprint and computation power. While the state-of-the-art machine learning models based on deep neural networks with millions of parameters can be easily deployed to a powerful workstation, they are not yet ready for the deployment on mobile devices having limited memory, computing power, and battery life. To reduce the storage and computation required for deployment on such devices, existing approaches include compressing a pretrained network by weight quantization, low-rank approximations and parameter pruning \cite{cheng2017survey} or designing a network topology with fewer parameters and computations \cite{howard2017mobilenets, tran2017improving}. It should be noted that most of these works focus on the convolutional architectures which are the core component of many visual learning systems. While visual inference tasks are actively investigated using convolutional neural networks, the potential of neural networks encompasses a much wider range of applications, ranging from health-care monitoring to smart sensors, which are traditionally solved with densely connected topologies \cite{haselsteiner2000using, qin1997neural, antsaklis1990neural}. In this regard, learning a problem-dependent, compact network configuration makes a step towards the realization of inference systems on low-cost devices.

In this work, we focus on the problem of learning efficient and compact network structures by learning fully heterogeneous multilayer networks using GOPs. We propose an algorithm to progressively grow the network structures, both in width and depth, while searching for the optimal operator set, the synaptic weights of the newly added neurons and the corresponding decision function. The contributions of our work can be summarized as follows:
\begin{itemize}
\item We analyze the current drawbacks of the related algorithms and propose an efficient algorithm to overcome the existing shortcomings by learning data-dependent, fully heterogeneous multilayer architectures employing GOPs. The resulting networks can achieve a high degree of self-organization with respect to the operator set of each neuron, the number of neurons per layer and the number of layers.

\item In addition to the proposed algorithm, we also present three other variants that can be seen as simplified versions of it. This is followed by the discussion focusing on the advantages and possible limitations of each variant in comparison with our proposed algorithm.

\item To validate the efficiency of the proposed algorithm in comparison with other variants and existing progressive learning systems, we have conducted extensive experimental benchmarks on a variety of real-world classification problems.

\item We publish our implementation of all evaluated methods in this paper to facilitate future research in this area \footnote{https://github.com/viebboy/HeMLGOP}.
\end{itemize}

The remainder of the paper is organized as follows: In Section 2, we review the GOP model with POP, the progressive learning algorithm proposed in \cite{kiranyaz2017progressive}. Section 3 starts by discussing the shortcomings of POP and proceeds to present our approach towards the design of fully heterogeneous networks using GOPs. The discussion of other variants of our approach is also presented in Section 3. In Section 4, we provide details of our experiment protocols and quantitative analysis of the experiment results. Section 5 concludes our work.

\section{Related Work}
This Section describes the GOP model and the corresponding algorithm POP, proposed in \cite{kiranyaz2017progressive} to learn GOP-based networks.

\subsection{Generalized Operational Perceptron}
The neuronal computation of the $i$-th GOP neuron at layer $l+1$ is illustrated in Figure \ref{f1}. As mentioned in the previous section, each GOP is characterized by the adjustable synaptic weights $w^{l+1}_{ki}$ ($k=1,\dots, N_l$), the bias term $b^{l+1}_{i}$ and an operator set (nodal operator $\boldsymbol{\psi}^{l+1}_i$, pooling operator $\boldsymbol{\rho}^{l+1}_i$, activation operator $\boldsymbol{f}^{l+1}_i$). The form of each operator is selected from a predefined library of operators: $\boldsymbol{\psi}^{l+1}_{i} \in \mathbf{\Psi}$, $\boldsymbol{\rho}^{l+1}_{i} \in \mathbf{P}$, $\boldsymbol{f}^{l+1}_{i} \in \mathbf{F}$. The task of learning a network using GOP is, therefore, the search for the optimal operator set for each GOP and the corresponding synaptic weights. Given ($\boldsymbol{\psi}^{l+1}_{i}$, $\boldsymbol{\rho}^{l+1}_{i}$, $\boldsymbol{f}^{l+1}_{i}$), $b^{l+1}_{i}$ and $w^{l+1}_{ki}$ ($k=1,\dots, N_l$), the activities of the $i$-th neuron can be described by the following equations:

\begin{align}
z^{l+1}_{ki} &=  \boldsymbol{\psi}^{l+1}_{i}(y^{l}_{k}, w^{l+1}_{ki})\label{eq1} \\
x^{l+1}_{i} &= \boldsymbol{\rho}^{l+1}_{i}(z^{l+1}_{1i}, \dots, z^{l+1}_{N_{l}i}) + b^{l+1}_i\label{eq2} \\
y^{l+1}_{i} &= \boldsymbol{f}^{l+1}_{i}(x^{l+1}_i)\label{eq3}
\end{align}

In Eq. (\ref{eq1}), the nodal operator takes as input the $k$-th output of the previous layer and the synaptic weight $w^{l+1}_{ki}$ which connects the $k$-th neuron at layer $l$ to the $i$-th neuron at layer $l+1$. After applying function $\boldsymbol{\psi}^{l+1}_i$, the nodal operation produces a scalar $z^{l+1}_{ki}$.

In Eq. (\ref{eq2}), the outputs of the nodal operation over neurons from the previous layer are then combined through the pooling operation, which applies the pooling function $\boldsymbol{\rho}^{l+1}_{i}$ over $z^{l+1}_{ki}$ ($k=1,\dots, N_{l}$). The pooled result is then shifted by the bias term  $b^{l+1}_i$ to produce $x^{l+1}_i$.

In Eq. (\ref{eq3}), the output of the pooling operation then goes through the activation function $\boldsymbol{f}^{l+1}_i$ to produce the output $y^{l+1}_i$ of the $i$-th GOP in layer $l+1$.

An example of a library of operators, which is also used in our experiments, is shown in Table \ref{t2}. It is clear that when the operator set of a GOP is selected as (\textit{multiplication}, \textit{summation}, \textit{sigmoid}) then it operates as a conventional perceptron.

\begin{table}[]
	\begin{center}
		\caption{Operator set library}\label{t2}
		\resizebox{0.9\linewidth}{!}{
			\begin{tabular}{|c|c|}\hline
				
				\textbf{Nodal} ($\mathbf{\Psi}$)  		& \textbf{$\boldsymbol{\psi}^{l+1}_{i}(y^{l}_{k}, w^{l+1}_{ki})$} \\ \hline \hline
				Multiplication			  		& $w_{ki}^{l+1} y_k^l$ \\ \hline
				Exponential						& $\exp (w_{ki}^{l+1}y_k^l)-1$ \\ \hline
				Harmonic						& $\sin(w_{ki}^{l+1}y_k^l)$ \\ \hline
				Quadratic						& $w_{ki}^{l+1}(y_k^l)^2$ \\ \hline
				Gaussian						& $w_{ki}^{l+1} \exp(-w_{ki}^{l+1}(y_k^l)^2)$ \\ \hline
				DoG								& $w_{ki}^{l+1}y_k^l \exp(-w_{ki}^{l+1}(y_k^l)^2)$ \\ \hline \hline
				
				\textbf{Pool} ($\mathbf{P}$)		& $\boldsymbol{\rho}^{l+1}_{i}(z^{l+1}_{1i}, \dots, z^{l+1}_{N_{l}i})$ \\ \hline
				Summation 						& $\sum_{k=1}^{N_l} z_{ki}^{l+1}$ \\ \hline
				1-Correlation 					& $\sum_{k=1}^{N_l-1} z_{ki}^{l+1}z_{(k+1)i}^{l+1}$ \\ \hline
				2-Correlation 					& $\sum_{k=1}^{N_l-2} z_{ki}^{l+1}z_{(k+1)i}^{l+1}z_{(k+2)i}^{l+1}$ \\ \hline
				Maximum 						& $\underset{k}\max (z_{ki}^{l+1})$ \\ \hline \hline
				
				\textbf{Activation} ($\mathbf{F}$)		& $\boldsymbol{f}^{l+1}_{i}(x^{l+1}_i)$ \\ \hline
				Sigmoid					& $1 / (1+\exp(-x^{l+1}_i))$	\\ \hline
				Tanh					& $\sinh(x^{l+1}_i) /\cosh(x^{l+1}_i)$ \\ \hline
				ReLU					& $\max(0,x^{l+1}_i)$ \\ \hline
				Softplus				& $\log(1+\exp(-x^{l+1}_i))$ \\ \hline
				Inverse Absolute		& $x^{l+1}_i /(1+|x^{l+1}_i|)$ \\ \hline
				ELU						& $x^{l+1}_i \mathbf{1}_{x^{l+1}_i \geq 0} + \exp(x^{l+1})_i \mathbf{1}_{x^{l+1}_i < 0}$ \\ \hline
				
			\end{tabular}
		}
	\end{center}
\end{table}

\subsection{Progressive Operational Perceptron}\label{POP}
Let $N_{\mathbf{\Psi}}$, $N_{\mathbf{P}}$, $N_{\mathbf{F}}$ be the number of elements in $\mathbf{\Psi}$, $\mathbf{P}$ and $\mathbf{F}$ respectively, then the total number of possible combinations for a GOP is $N_\mathbf{O} = N_{\mathbf{\Psi}}*N_{\mathbf{P}}*N_{\mathbf{F}}$. It is clear that given a multilayer topology of GOPs and a large value for $N_\mathbf{O}$, the combinatorial search space when optimizing all neurons simultaneously in such network is enormous. For example, consider the case where $N_\mathbf{O}=72$, as used in \cite{kiranyaz2017progressive}, and using a two-hidden-layer network with $100$ neurons in each layer and $10$ output neurons. Such a small network architecture corresponds to $72^{210}$ different configurations. To narrow the search space, POP was proposed in \cite{kiranyaz2017progressive} to learn the network topology in a layerwise manner. In order to operate, a template network structure specifying the number of neurons for each hidden layer and maximum number of hidden layers is pre-defined. In addition, a target objective is specified to determine the convergence of the algorithm. For example, [$N_i$, $N_1$, $N_2$, $N_3$, $N_o$; $mse=\epsilon$] defines a template with $N_i$ input units, $N_o$ output neurons, 3 hidden layers with $N_1$, $N_2$, $N_3$ neurons respectively, and $\epsilon$ specifies the target mean squared error. Starting from the first hidden layer, POP constructs a Single Hidden Layer Network (SHLN) [$N_i$, $N_1$, $N_o$] and learns the operator sets and weights in the hidden and output layer with a constraint: neurons in the same layer share the same operator set. Finding the operator sets is done by a greedy iterative search procedure called \textit{two-pass GIS}. In the first pass, a random operator set $\phi_h$ is fixed to the hidden layer and POP iterates through all operator sets in the library for the output layer: at each iteration, the output layer is assigned the iterated operator set $\phi_o$; the synaptic weights of SHLN with ($\phi_h$, $\phi_o$) are found by BP for $E$ epochs, and the performance is recorded. After this procedure, the current best operator set $\phi^{*}_o$ in the output layer is found. With $\phi^{*}_o$ fixed in the output layer, POP performs similar loop to find the best set for the hidden layer $\phi^{*}_{h}$. The second pass of GIS is similar to the first one, except the operator set in the hidden layer is initialized with $\phi^{*}_{h}$ from the first pass instead of random assignment. After applying two-pass GIS, the performance of the current SHLN is compared with the target objective $\epsilon$. If the target is not achieved, the output layer of SHLN is discarded and the current hidden layer is fixed and used to generate fixed inputs to learn the next hidden layer with $N_2$ neurons in the similar manner as the first hidden layer. On the other hand, if the target objective is met, POP stops the progressive learning procedure and fine-tunes all the weights and biases in the network structure that has been learned.

\subsection{Limitations in POP}\label{pop_limitations}
It is clear from Section \ref{POP} that POP optimizes the operator set and weights in each hidden layer by running through $4$ loops over the library of the operator sets with each iteration running BP with $E$ epochs. Therefore, the computational complexity to optimize an SHLN is $O(4N_\mathbf{O}E)$ BP epochs. Such a search scheme is not only computationally demanding, but also redundant due to the fact that when the target objective cannot be achieved with the current network configuration, POP simply discards what has been learned for the output layer and reiterates the searching procedure for the new hidden and output layer. Let us consider the case where the operator set in the output layer is already known a priori, the cost of the search in POP is reduced from $O(4N_\mathbf{O}E)$ to $O(N_\mathbf{O}E)$, which is a significant factor of reduction. In fact, we argue that if the hidden neurons can extract highly discriminant representations, which is the design target of GOP, then only a simple linear decision function is needed in the output layer.

There are two constraints imposed by POP on the learned architecture: a predefined width of each hidden layer and the sharing of the operator set within the same layer. Both constraints limit the representational power of the learned hidden layer. While it is computationally infeasible to search for the operator set of each individual GOP following the searching approach in POP, \cite{kiranyaz2017progressive} argued that in a classification problem an optimal operator set of a neuron should also be optimal to others in the same layer, i.e., on the same level of the hierarchical abstract. This is, however, a strong assumption. As an illustrational example, let us assume that, at some arbitrary level of abstraction in the network, there appears patterns of both straight lines and curves, and we assume that there exist two operator sets that allow the neurons to detect straight lines and curves respectively. By limiting the neurons to either being able to detect a line or a curve, whichever yields better results, one of the patterns will not be captured in the internal representation of the network. Such an approach will lead to the confusion on the objects which are composed of both patterns. One might argue that with enough neurons that can detect lines, a curvature can also be detected in a limiting sense. This comes to the question: \textit{how many neurons will be enough?}. By imposing a predefined width of each hidden layer, POP already incurs an additional hyper-parameter choice, leading to either insufficient or redundant representation, both of which require a huge amount of hyper-parameter tuning efforts to achieve an efficient and compact network.

\section{Proposed Method}\label{proposed_method}
In this section, we describe a new approach to overcome the limitations of current algorithms in building heterogeneous network architectures directly from data. At the end of this section, we also discuss other possible variants of our approach and present our view on the pros and cons of each of them.

\subsection{Heterogeneous Multilayer Generalized Operational Perceptron (HeMLGOP)}

The aforementioned limitations in POP are, in fact, inter-related. The computational complexity of the search procedure can be reduced by avoiding the search of the operator sets in the hidden layer in conjunction with the output layer. This can be done by simply assuming a linear decision function, which requires highly discriminative hidden representations. In order to produce highly discriminative hidden representations, heterogeneous hidden layers of GOPs with adaptive size might be desirable. It should be noted that, in order to search for the optimal operator set of a neuron, it is necessary to evaluate all operator sets in the library. Instead of evaluating each operator set by hundreds BP iterations as in POP, we propose to use ideas originated from Randomized Networks (RN) \cite{pao1992functional, schmidt1992feedforward, huang2006extreme, rahimi2008random} for the evaluation of an operator set. Given a single hidden layer topology with linear transformation in the output layer, we assign random synaptic weights connecting the input layer to the hidden layer while giving a closed-form global solution of the output layer, i.e., the decision function. In particular, let $\mathbf{H} \in \mathbb{R}^{N\times d}$ and $\mathbf{Y} \in \mathbb{R}^{N\times C}$ be the hidden representation, and the target representation of $N$ training samples respectively. The optimal synaptic weights $\mathbf{B} \in \mathbb{R}^{d \times C}$ connecting the hidden layer and the output layer is given as:

\begin{equation}\label{eq4}
\mathbf{B} = \mathbf{H}^{\dagger}\mathbf{Y}
\end{equation}
where $\mathbf{H}^{\dagger}$ is the Moore-Penrose generalized inverse of $\mathbf{H}$.

There are several methods to calculate the Moore-Penrose generalized inverse of a matrix \cite{serre2002matrices, banerjee1973generalized}. For example, in the orthogonal projection method, $\mathbf{H}^{\dagger} = \mathbf{H}^{T}(\mathbf{H}\mathbf{H}^T + c\mathbf{I})^{-1}$ when $d>N$ or  $\mathbf{H}^{\dagger} = (\mathbf{H}^T\mathbf{H} + c\mathbf{I})^{-1} \mathbf{H}^T$ when $d<N$, with $c$ being a positive scalar added to the diagonal of $\mathbf{H}^T \mathbf{H}$ or $\mathbf{H} \mathbf{H}^T$ to ensure stability and improve generalization performance according to ridge regression theory \cite{hoerl1970ridge}.

Given a single hidden layer network with GOP neurons in the hidden layer and linear output layer, and since each operator set represents a distinct type of neuronal activity or functional form, we argue that the suitable functional form of a GOP, i.e., the operator set, can be evaluated with random synaptic weights drawn from a uniform distribution \cite{liu2015extreme}. After finding the optimal operator set of a neuron with respect to the objective function, the corresponding weights can be fine-tuned by BP.

To learn a problem-dependent network topology, we adopt a progressive learning approach in terms of width and depth in our algorithm. Given a learning problem, the proposed algorithm starts with a single hidden layer network of $n_{\min}$ GOPs in the hidden layer and $C$ linear neurons in the output layer. By starting with a small $n_{\min}$, e.g., $n_{\min}=2$, we assume these GOPs share the same operator set. The algorithm proceeds to select the optimal operator set of $n_{\min}$ neurons by iterating through all operator sets in the library: at iteration $j$, random synaptic weights in the hidden layer are drawn from a uniform distribution, the decision boundary is calculated as follows:

\begin{equation}\label{eq5}
\mathbf{B}_{n_{\min}}^{j} = \bar{\mathbf{H}}_{n_{\min}}^{j\dagger}\mathbf{Y}
\end{equation}
where $\bar{\mathbf{H}}_{n_{\min}}^{j}$ denotes the standardized hidden output of $n_{\min}$ GOPs with operator set $\phi_j$.

At each iteration, the performance of the network is recorded. After evaluating all $N_\mathbf{O}$ operator sets, the best performing one $\phi_{n_{\min}}^{*}$ is selected for the current $n_{\min}$ neurons and the corresponding synaptic weights $\mathbf{W}_{n_{\min}}^{*}$ as well as output layer weights $\mathbf{B}_{n_{\min}}^{*}$ are updated with BP for $E$ epochs. During BP with mini-batch training, the normalization step is replaced by Batch Normalization \cite{ioffe2015batch}, which is initialized with mean and standard deviation. Since the hidden layer will be incrementally grown with heterogeneous GOPs, the normalization step is necessary to ensure that the hidden representations in the network have similar range. Once the operator set is found and the synaptic weights of a GOP are fine-tuned with BP, they are fixed.

The algorithm continues by progressively adding $n_i$ GOPs sharing the same operator set to the hidden layer. It is worth noting that when $n_i=1$, the algorithm allows the growth of fully heterogeneous layers. The operator set of $n_i$ newly added GOPs is found similarly as in case of the first $n_{\min}$ GOPs, i.e., by iterating through all operator sets and solving for the output layer weights. In particular, at iteration $j$, let $\bar{\mathbf{H}}_c^{*}$ be the normalized hidden representation of the existing GOPs that have been learned and $\bar{\mathbf{H}}_{n_i}^{j}$ be the normalized hidden representation produced by the newly added GOPs with the $j$-th operator set in the library and random weights, then the optimal linear transformation in the output layer is given as:

\begin{equation}\label{eq6}
\mathbf{B}_{c+i}^{j} = [\bar{\mathbf{H}}_c^{*}, \bar{\mathbf{H}}_{n_i}^j]^{\dagger} \mathbf{Y}
\end{equation}

As an alternative to Eq. (\ref{eq6}), the new decision boundary $\mathbf{B}_{c+i}^{j}$ can also be updated efficiently in an incremental manner based on $\mathbf{B}_{c}^{*}$, the decision boundary with respect to $\bar{\mathbf{H}}_c^{*}$ \cite{courrieu2008fast}. After the best performing operator set $\phi_{n_i}^{*}$ of $n_i$ newly added GOPs is found, their synaptic weights $\mathbf{W}_{n_{i}}^{*}$, the normalization statistics and the linear transformation $\mathbf{B}_{c+i}^{*}$ in the output layer are updated through BP. Here we should note that the existing GOPs with $\mathbf{H}_c^{*}$ representation are not updated since the inclusion of $n_i$ neurons is expected to complement the existing features. While the update of all the neurons, including the existing GOPs, might produce better performance, it can also lead the network to over-fitting regime by forcing the co-adaptation of all neurons to the training data. Thus, by only updating the weights and biases of the newly added GOPs, we also enforce a form of regularization. The progressive learning in the current hidden layer stops when the inclusion of new neurons stops improving the performance of the network. This is measured by a relative criterion based on the rate of improvement rather than an absolute threshold value on the performance as in POP:

\begin{equation}\label{eq7}
r_{i} = \frac{\mathcal{L}_c - \mathcal{L}_{c+i}}{\mathcal{L}_{c}}
\end{equation}
where $r_i$ denotes the rate of improvement when adding $n_i$ neurons to the current hidden layer and $\mathcal{L}_c$, $\mathcal{L}_{c+i}$ denote the loss values before and after adding neurons respectively. For large positive $r_i$, the inclusion of new neurons indicates a large improvement of the network with respect to the existing structure. On the contrary, a small or negative $r_i$ indicates a minimal improvement or a degradation in the performance. It should be noted that depending on the learning problem, other quantities can be chosen in place of the loss value, with appropriate signs in the numerator of Eq. (\ref{eq7}). For example, in classification problem, $r_i$ can be defined as $(\mathcal{A}_{c+i} - \mathcal{A}_{c})/\mathcal{A}_c$ with $\mathcal{A}$ denotes the classification accuracy. Given $r_i$ and a threshold $\epsilon_{n}$, the proposed algorithm stops adding neurons to the current hidden layer if $r_i < \epsilon_{n}$.

When the progression in the current hidden layer stops improving performance with respect to $\epsilon_{n}$, the proposed algorithm forms a new hidden layer with $n_{\min}$ GOPs between the current hidden layer and the output layer. All the existing hidden layers in the network are then fixed and act as feature extractor, producing inputs to the newly formed hidden layer. The progression in the new hidden layer repeats in a similar manner as in the previous hidden layers with an initial $n_{\min}$ GOPs and incrementally adding $n_i$ GOPs until the criterion $r_i < \epsilon_{n}$ is met. After the new hidden layer is fully learned, the proposed algorithm evaluates the necessity to include the newly learned hidden layer by evaluating the relative improvement of the network before and after adding the new hidden layer:

\begin{algorithm}
	\caption{Heterogeneous Multilayer Generalized Operational Perceptron (HeMLGOP)}
	\label{HeMLGOP}
	\begin{algorithmic}[1]
		\State \textbf{Inputs}: $\mathbf{X} \in \mathbb{R}^{N\times D}$, $\mathbf{Y} \in \mathbb{R}^{N\times C}$, $\mathbf{\Phi} = \mathbf{\Psi} \Cross \mathbf{P} \Cross \mathbf{F}$, $n_{\min}$, $n_i$, $\epsilon_n$, $\epsilon_l$.
		
		\State \textbf{Initialization}: $\Phi^{*} = \{\}$, $\mathbf{W}^{*} = \{ \}, l=1 $.
		\While{True}
		\State For $\phi_j \in \mathbf{\Phi}$: calculate $\mathbf{B}_{n_{\min}}^{j}$ as in Eq. (\ref{eq5}).
		\State Select $\phi_{n_{\min}}^{*}$, $\mathbf{B}_{n_{\min}}^{*}$, $\mathbf{W}_{n_{\min}}^{*}$.
		\State Fine-tune $\mathbf{B}_{n_{\min}}^{*}$, $\mathbf{W}_{n_{\min}}^{*}$.
		\State Update $\bar{\mathbf{H}}_c^{*} = \bar{\mathbf{H}}_{n_{\min}}^{*}$, $\bar{\mathbf{B}}_c^{*} = \bar{\mathbf{B}}_{n_{\min}}^{*}$
		\State $\Phi_l^{*} = \{ \phi_{n_{\min}}^{*} \}$ and $\mathbf{W}_{l}^{*} =\{ \mathbf{W}_{n_{\min}}^{*} \}$.
		
		\While{True}
		\State For $\phi_j \in \mathbf{\Phi}$: calculate $\mathbf{B}_{c+i}^{j}$ as in Eq. (\ref{eq6}).
		\State Select $\phi_{n_{i}}^{*}$, $\mathbf{B}_{c+i}^{*}$, $\mathbf{W}_{n_{i}}^{*}$.
		\State Fine-tune $\mathbf{B}_{c+i}^{*}$, $\mathbf{W}_{n_{i}}^{*}$.
		\State Calculate $r_i$ as in Eq. (\ref{eq7}).
		\If{$r_i < \epsilon_n}$
		\State \textbf{break}
		\Else
		\State Update $\bar{\mathbf{H}}_c^{*} = [\bar{\mathbf{H}}_c^{*}, \bar{\mathbf{H}}_{n_i}^{*}]$, $\bar{\mathbf{B}}_c^{*} = \bar{\mathbf{B}}_{c+i}^{*}$
		\State $\Phi_l^{*} = \Phi_l^{*} + \{ \phi_{n_i}^{*} \}$ and $\mathbf{W}_l^{*} = \mathbf{W}_l^{*}  + \{\mathbf{W}_{n_i}^{*}\}$.
		\EndIf
		\EndWhile
		\State Calculate $r_l$ as in Eq. (\ref{eq8}).
		\If{$r_l < \epsilon_l$}
		\State \textbf{break}
		\Else
		\State Update $\mathbf{X} = \bar{\mathbf{H}}_c^{*}$, $\Phi^{*} = \Phi^{*} + \{ \Phi_l^{*} \}$
		\State and $\mathbf{W}^{*} = \mathbf{W}^{*} + \{ \mathbf{W}_l^{*} \}$.
		\EndIf
		\State Update $l=l+1$
		\EndWhile
		\State Fine-tune $\mathbf{W}^{*}$, $\mathbf{B}_c^{*}$.
		\State \textbf{Outputs}: $\Phi^{*}$, $\mathbf{W}^{*}$, $\mathbf{B}_c^{*}$
	\end{algorithmic}
\end{algorithm}

\begin{equation}\label{eq8}
r_{l+1} = \frac{\mathcal{L}_{l} - \mathcal{L}_{l+1}}{\mathcal{L}_{l}}
\end{equation}
where $r_{l+1}$ denotes the rate of improvement when adding the new hidden layer and $\mathcal{L}_{l}$, $\mathcal{L}_{l+1}$ denote the loss values before and after adding the new hidden layer respectively. Similar to the progression of neurons in a hidden layer, the progression of hidden layers is controlled through hyper-parameter $\epsilon_{l}$. The newly learned hidden layer is included in the network topology and the progression continues if $r_{l+1} \geq \epsilon_{l}$. Otherwise, the progressive learning terminates. After that, all the synaptic weights and biases of the network are fine-tuned through Back Propagation. On one hand, the final fine-tuning step allows the co-adaption of all neurons in the network to fit the training data, which might lead the network to the over-fitting regime. On the other hand, if the network produced by progressive learning step under-fits the problem, the fine-tuning step allows the network to better fit the problem. Thus, the necessity of this fine-tuning step is problem-dependent and is evaluated based on the performance on the training set (or validation set if exists). That is, if the training (validation) performance improves after the fine-tuning step, the fine-tuned network is used, otherwise, the network learned by the progressive learning step is used. Since a new hidden layer is initially formed with a small number of neurons, our proposed algorithm only evaluates the inclusion of a new hidden layer when it is fully learned. The summary of our proposed algorithm is presented in Algorithm \ref{HeMLGOP}.

\subsection{HeMLGOP Variants}\label{proposed_discussion}
One can also identify three variants of the proposed algorithm. They are the following:

\begin{itemize}
\item \textit{Homogeneous Multilayer Randomized Network (HoMLRN)}: instead of a heterogeneous layer of GOPs, in this variant, all neurons in the same layer share the same operator set. A hidden layer starts with $n_{\min}$ neurons whose operator set is found by using RN in the operator set search procedure. Once the operator set of the starting $n_{\min}$ neurons is found, it is fixed for all other neurons in the same layer. At each progressive step, $n_i$ neurons with random weights are added to the current hidden layer, and the decision boundary is adjusted through linear regression. After the progression finishes, the final network structure is fine-tuned through BP.

\item \textit{Heterogeneous Multilayer Randomized Network (HeMLRN)}: in this variant, the progressive learning procedure is similar to our algorithm, i.e., for newly added neurons, the algorithm searches for the best performing operator set by using RN. The only difference between this variant and our algorithm is that the synaptic weights are not fine-tuned during progressive learning but only after the final topology is found.

\item \textit{Homogeneous Multilayer Generalized Operational Perceptron (HoMLGOP)}: similar to HoMLRN and POP, this variant enforces the sharing of operator set within the same layer. The progressive learning in a hidden layer starts with $n_{\min}$ GOPs whose operator set is found via Randomized Network in the operator set search procedure. GOPs with the operator set the same as the starting $n_{\min}$ neurons are incrementally added to the current hidden layer. At each increment, the synaptic weights of newly added neurons are updated through BP instead of Randomized Network as in HoMLRN. After the progression of network structure, the final topology is fine-tuned as a whole.
\end{itemize}

It is clear that the aforementioned variants can be seen as simplified versions of our approach in certain aspects. Particularly, both HoMLRN and HeMLRN depend solely on Randomized Networks during progressive learning, which reduces a portion of computational cost induced by weights finetuning through BP. While Randomized Networks can be suitable to search for the functional form of newly added GOPs, we argue that it is necessary to further adjust the weights of the newly added neurons through BP to effectively exploit their representation power. Without the weight adjustment step interleaved with Randomized Network, both HoMLRN and HeMLRN are expected to progress towards having large hidden layers. Moreover, since the hidden layers rely only on random weights during the progression, the outputs of a hidden layer are not expected to be highly discriminative as an input to the next hidden layer, which might also lead to ineffective progression in depth. While HoMLGOP incorporates the weight finetuning step in the progressive learning procedure, this variant avoids the cost of searching for the optimal operator set when incrementally adding neurons. As a result, the homogeneity constraint in hidden layers might prevent HoMLGOP from learning compact hidden layers.

\subsection{Convergence analysis}
The proposed algorithm consists of two processing phases:
\begin{itemize}
    \item \textit{Network architecture initialization}: During the progressive learning phase, the network’s architecture is determined in a layer-wise manner. This phase not only determines the structure of the network, i.e. the number of layers and number of neurons per layer, but also determines the characteristics of each neuron (or each added block of neurons). During this stage, the progression in each layer converges, i.e., adding new blocks to the same layer produces a converging sequence of the loss values (MSE). The proof is given in the Appendix \ref{A2} and we also provide illustrations of the training curves obtained in our experiments in Appendix \ref{A3}. While each of the steps in phase one converges, there is no guarantee that adding new hidden layers will produce a monotonically decreasing sequence of the training losses considering the loss values obtained for different layers of the network. Since the output layer is a linear layer which linearly combines the dimensions of the data representations of the previous layer, to ensure this property, we must have a mechanism to ensure that the subspace spanned by the representations from the newly added layer overlaps with the subspace spanned by the training targets more than the previous hidden layer does. In order to develop this property, more research is required, which would be a very interesting and challenging future research work.

    \item \textit{Combined network training}: This phase corresponds to the training (fine-tuning in our case) of a feedforward network the structure of which and the initial parameters are obtained by applying the network architecture initialization process of the first case. Since all layers are formed by neurons with well-behaved nodal, pooling and activation operators (in the sense of differentiation), this optimization process converges to a local optimum following stochastic gradient descent \cite{robbins1985stochastic}.
\end{itemize}
Thus, we can conclude that the first phase leads to a network structure the parameters of which are initialized to a set of parameters achieving a good loss value and are further optimized through BP to reach a local minimum of the loss function.

It is worth mentioning here that the strategy followed for stopping the network architecture initialization phase is important for the generalization performance of the resulting network. As mentioned above, when the growth of neurons in the last hidden layer converges, HeMLGOP forms a new hidden layer. New blocks are added to this newly formed layer until convergence. HeMLGOP then evaluates whether the addition of this new hidden layer improves performance. Without any constraint, there is no guarantee that adding a new hidden layer will improve the training loss. When the addition of a new hidden layer produces a smaller training MSE value compared to the previous hidden layer, it only indicates that the new features produced by the new hidden layer span a feature space that can better approximate the training targets in the MSE sense. For function approximation tasks, this property is desirable. However, for prediction tasks that require generalization, improvement of the training loss does not guarantee the improvement of the generalization of the model to unseen data. With the availability of the validation data, improvement obtained by the addition of the new hidden layer can be evaluated based on the performance on the validation set. An improvement on the validation set performance means that the feature space determined by the outputs of the newly added layer can better approximate the validation targets and, thus, incorporating the new layer in the network increases the generalization ability of the model.

\subsection{Complexity analysis}
Since the proposed algorithm is a progressive algorithm, its computational complexity is the sum of the computational complexity of all progressive steps, each depending on the current network specifications, i.e. the dimension of the current and previous hidden layer, the operator set of each GOP neuron. We provide the computational complexity estimation of adding a new block of neurons, given the current network settings in Appendix \ref{A1}.

It is not straightforward to compare the computational complexity of the proposed algorithm and its variants, POP and related algorithms, since the complexity of all progressive learning algorithms depends on the speed of convergence, the number of hidden layers added and the selected operators. However, it is worth noting that since the proposed algorithm allows the growth of heterogeneous hidden layers, and at each incremental step, the synaptic weights of newly added neurons are strengthened through BP to fully adapt to the problem, we expect to observe the rate of improvement to converge quickly, producing both compact and efficient network structures. In our empirical analysis, the proposed algorithm converges after a few incremental steps in most of the learning problems, and the number of network parameters in the learned architectures is much lower compared to the competing approaches. Another point worth mentioning is that the search procedure in the proposed method and the mentioned variants relies on random weights, which might produce different operator sets at different runs. This can lead to high variance in the final topologies between different runs, especially in case of HoMLRN and HoMLGOP in which the operator set of a layer is found only once. The effect of randomness is, however, reduced in the proposed algorithm because in the case where an optimal operator set was not found in the previous incremental steps due to randomness, it can still be chosen in the next steps.

\section{Experiments}

In this section, we provide empirical analysis of the proposed algorithm, the aforementioned variants, and other related algorithms. We start by describing the experimental protocols and datasets, followed by the discussion of the empirical results.

\subsection{Competing methods}
In order to compare the effectiveness of the proposed algorithm with related approaches, the following additional methods were included in our empirical analysis:

\begin{itemize}
\item Progressive Operational Perceptron (POP) \cite{kiranyaz2017progressive}: the only existing GOP-based algorithm.

\item HoMLRN, HeMLRN, HoMLGOP: the $3$ variants of the proposed method mentioned in the previous section.

\item Progressive Learning Network (PLN) \cite{chatterjee2017progressive}: by using nonlinear activation functions that satisfy the progression property, the authors in \cite{chatterjee2017progressive} proposed a progressive learning algorithm that increments the width of a hidden layer by random perceptrons and solving a convex optimization problem. When the performance improvement in a hidden layer saturates, PLN forms a new hidden layer by incrementally adding random neurons to the current output layer to form a new hidden layer and adding new output layer.

\item Broad Learning System (BLS) \cite{chen2018broad}: based on the idea of Random Vector Functional Link Neural Network \cite{pao1994learning}, the authors proposed a progressive learning algorithm that increments the width of a two hidden layer network. Neurons in the first hidden layer are called feature nodes, which synthesize hidden features by random linear transformation and sigmoid activation. Neurons in the second hidden layer are called enhancement nodes, which again linearly transform the outputs of feature nodes with random weights, followed by the sigmoid activation. The outputs of feature nodes and enhancement nodes are concatenated as an input to a linear output layer. Before progressively adding new feature nodes and enhancement nodes, BLS fine-tunes the feature nodes by Alternating Direction Method of Multiplier (ADMM). During the progression, only random nodes are added.

\item Progressive Multilayer Perceptron (PMLP): this is a variant of POP that uses McCulloch-Pitts perceptron instead of GOP. The progressive learning step is similar to POP with a pre-defined maximal template structure as an input, PMLP incrementally adds a new hidden layer if a target objective cannot be achieved.
\end{itemize}

\subsection{Datasets}
We have conducted experiments on $11$ classification problems in different application domains with different sizes, ranging from few hundred samples up to $60$k samples. With respect to the problem size, the $11$ datasets can be divided into $2$ groups: small-scale problems ($5$ datasets) formed by less than $2000$ samples and medium/large scale problems ($6$ datasets) formed by more than $2000$ samples.

Information about all datasets used in our experiments is presented in Table \ref{t1}. For PIMA \cite{smith1988using}, CMC \cite{niebles2010modeling} and YEAST \cite{horton1996probabilistic}, we used the original data representations provided by the database. Olympic Sports \cite{niebles2010modeling} and Holywood3d \cite{hadfield2013hollywood} are human action video datasets. We used the state-of-the-art action video representation in \cite{wang2013action} and combined the five action descriptions following the suggested multi-channel kernel approach followed by KPCA to obtain vector-based representation for each action video. All medium/large scale problems are classification problem based on visual inputs. Particularly, 15 scenes and MIT indoor are scene recognition datasets, Caltech101 and Caltech256 are related to the problem of object recognition while CFW and PubFig are face recognition problems. Regarding the input representation of scene recognition and object recognition datasets, we employed the deep features generated by average pooling over spatial dimension of the last convolution layer of VGG network \cite{simonyan2014very} trained on ILSVRC2012 database. Similarly, deep features generated by VGGface network \cite{parkhi2015deep} were used in CFW and PubFig.

\begin{table}[t]
	\begin{center}
		\caption{Datasets Information}\label{t1}
		\resizebox{\linewidth}{!}{
			\begin{tabular}{|c|c|c|c|}\hline
				
				Database 			& $\#$Samples			& Input dimension			& Target dimension 		\\ \hline \hline
		
				PIMA \cite{smith1988using}				& 768					& 8							& 2						\\ \hline	
				Olympic Sports \cite{niebles2010modeling}	& 774					& 100							& 16						\\ \hline
				Holywood3d \cite{hadfield2013hollywood}		& 945					& 100							& 14						\\ \hline		
				CMC	\cite{lim2000comparison}			& 1473					& 9							& 3						\\ \hline
				YEAST \cite{horton1996probabilistic}				& 1484					& 8							& 10						\\ \hline \hline

				15 scenes \cite{lazebnik2006beyond}			& 4485					& 512							& 15						\\ \hline	
				MIT indoor \cite{quattoni2009recognizing}		& 15620					& 512							& 67						\\ \hline	
				Caltech101 \cite{fei2007learning}		& 9145					& 512							& 102						\\ \hline	
				Caltech256 \cite{griffin2007caltech}		& 30607					& 512							& 257						\\ \hline	
				PubFig	\cite{pinto2011scaling}		& 13002					& 512							& 83						\\ \hline	
				CFW60k \cite{zhang2012finding}				& 60000					& 512							& 500						\\ \hline		
				
			\end{tabular}
		}
	\end{center}
\end{table}

Since POP is the most computationally demanding algorithm, we could only afford to perform experiments with POP in small-scale problems. Although empirical results in medium/large scale problems are not available, the efficiency of POP in comparison with other algorithms can be observed in five small-scale datasets.

\begin{table*}[t!]
	\begin{center}
		\caption{Classification accuracy (\%)}\label{t3}
		\resizebox{\textwidth}{!}{
			\begin{tabularx}{\textwidth}{ |c| *{8}{Y|} }\cline{2-9}
				\multicolumn{1}{c|}{}

& HeMLGOP 		& HoMLGOP		& HeMLRN			& HoMLRN 		& POP		& PMLP 		& PLN		& BLS	\\ \hline
Holywood3d 		& $78.78$ 	& $76.72$  & $78.93$   & $74.42$   & $74.42$   & $\mathbf{79.38}$   & $75.24$   & $71.38$  \\ \hline
Olympic Sports 	& $87.06$ 	& $87.08$  & $87.50$   & $83.92$   & $\mathbf{87.82}$  & $87.45$   & $87.23$   & $85.88$ \\ \hline
CMC 			& $60.37$ 	& $59.36$  & $\mathbf{62.63}$   & $58.90$   & $59.75$   & $59.95$   & $58.93$   & $55.23$  \\ \hline
PIMA  			& $\mathbf{81.81}$		& $79.84$	 & $81.41$   & $76.68$	& $79.05$		& $78.66$	& $71.94$		& $68.77$ \\ \hline
YEAST  			& $\mathbf{65.75}$		& $65.31$		& $65.53$		& $63.69$		& $64.30$		& $63.29$		& $63.08$ 	& $55.57$ \\ \hline \hline
15 scenes  		& $\mathbf{92.01}$		& $91.57$		& $91.80$		& $90.24$		& -			& $91.68$		& $89.46$		& 86.36 \\ \hline
MIT indoor  	& $\mathbf{69.77}$		& $69.14$		& $69.23$		& $67.99$		& - 		& $69.18$		& $64.38$		& $56.23$ \\ \hline
Caltech101  	& $\mathbf{92.08}$		& $91.41$		& $91.68$		& $90.87$		& - 		& $91.78$		& $89.64$		& $84.28$	\\ \hline
Caltech256  	& $79.21$		& $78.61$		& $79.09$		& $77.65$		& - 		& $\mathbf{79.26}$		& $75.58$		& $70.60$	\\ \hline
PubFig  		& $\mathbf{98.92}$		& $98.29$		& $98.85$		& $98.69$		& - 		& $98.85$		& $98.46$		& $95.00$	\\ \hline
CFW60k  			& $88.15$		& $\mathbf{88.23}$		& $87.90$		& $57.80$		& - 		& $87.93$		& $85.09$		& $75.19$ \\ \hline

			\end{tabularx}
		}
	\end{center}
\end{table*}

\begin{table*}[t!]
	\begin{center}
		\caption{Model sizes (\#parameters)}\label{t6}
		\resizebox{\textwidth}{!}{
			\begin{tabularx}{\textwidth}{ |c| *{8}{Y|} }\cline{2-9}
				\multicolumn{1}{c|}{}
				
& HeMLGOP 		& HoMLGOP		& HeMLRN			& HoMLRN 		& POP		& PMLP 		& PLN		& BLS	\\ \hline
Holywood3d 		& $9.4$k 	& $7.0$k  	& $14.0$k   & $18.7$k   & $64.0$k   & $23.4$k   & $\mathbf{6.4}$k   & $58.8$k  \\ \hline
Olympic Sports 	& $7.1$k 	& $11.9$k  	& $23.8$k   & $11.9$k   & $64.1$k  	& $105.0$k   & $\mathbf{6.8}$k   & $43.84$k \\ \hline
CMC 			& $\mathbf{0.9}$k 	& $\mathbf{0.9}$k  	& $3.0$k   	& $1.5$k   & $43.6$k   & $84.2$k   & $4.9$k   & $7.9$k  \\ \hline
PIMA  			& $\mathbf{0.8}$k	& $4.5$k	& $2.3$k   	& $15.8$k	& $43.2$k	& $43.2$k	& $21.5$k	& $7.4$k \\ \hline
YEAST  			& $\mathbf{1.3}$k	& $\mathbf{1.3}$k	& $4.2$k	& $10.3$k	& $44.8$k	& $44.8$k	& $219.7$k 	& $18.8$k \\ \hline \hline
15 scenes  		& $\mathbf{63.6}$k		& $110.9$k		& $106.0$k		& $106.0$k		& -			& $187.2$k		& $134.2$k		& $252.3$k \\ \hline
MIT indoor  	& $\mathbf{69.9}$k		& $81.5$k		& $116.5$k		& $116.5$k		& - 		& $116.5$k		& $122.3$k		& $513.8$k \\ \hline
Caltech101  	& $\mathbf{74.1}$k		& $98.8$k		& $123.5$k		& $123.5$k		& - 		& $204.7$k		& $166.9$k		& $653.0$k	\\ \hline
Caltech256  	& $\mathbf{92.9}$k		& $154.7$k		& $154.7$k		& $154.7$k		& - 		& $235.9$k		& $1466.2$k		& $2205.7$k	\\ \hline
PubFig  		& $\mathbf{47.9}$k		& $\mathbf{47.9}$k		& $119.7$k		& $119.7$k		& - 		& $160.3$k		& $130.8$k		& $591.1$k	\\ \hline
CFW60k 			& $\mathbf{162.9}$k		& $183.2$k		& $203.5$k		& $203.5$k		& - 		& $203.5$k		& $5472.6$k		& $5092.0$k \\ \hline

			\end{tabularx}
		}
	\end{center}
\end{table*}

\subsection{Experiment Protocol}

In small-scale problems, since the number of samples is small, we only partitioned the datasets into a train ($60\%$) and a test ($40\%$) set, except for Holywood3d and Olympic Sports in which we used the partition given by the databases. In medium/large scale problems, $60\%$ of the data was randomly chosen for training while $20\%$ was selected as validation and test set each. To deal with the effect of randomness, each algorithm was run $3$ times on each problem, and the median performance on the test set and the corresponding architectural information are reported.

Since other progressive learning methods (PLN, BLS) are significantly affected by hyper-parameter settings, we have conducted extensive hyper-parameter search for each algorithm. Particularly, in PLN, we tested the values $\lambda \in \{10^{-3},10^{-2},10^{-1},1,10^{1},10^{2},10^{3}\}$ for the least-square regularization parameter, and $\alpha \in \{10^{-3},10^{-2},10^{-1},1,10^{1},10^{2},10^{3}\}$ , $\mu \in \{10^{-3},10^{-2},10^{-1},1,10^{1},10^{2},10^{3}\}$ for the regularization parameters when solving the output layer; in BLS, the regularization parameter used in the calculation of pseudo-inverse is in the set of $\lambda \in \{10^{-3},10^{-2},10^{-1},1,10^{1},10^{2},10^{3}\}$ and regularization parameter used in ADMM algorithm is in the set of $\mu \in \{10^{-3},10^{-2},10^{-1},1,10^{1},10^{2},10^{3}\}$. The number of iterations in ADMM algorithm was set to $500$ for both PLN and BLS.

\begin{table*}[t!]
	\begin{center}
		\caption{Training time (in seconds) on small-scale problems on CPU}\label{t4}
		\resizebox{\textwidth}{!}{
			\begin{tabularx}{\textwidth}{ |c| *{8}{Y|} }\cline{2-9}
				\multicolumn{1}{c|}{}
				
				& HeMLGOP 		& HoMLGOP		& HeMLRN			& HoMLRN 		& POP		& PMLP 		& PLN		& BLS	\\ \hline
				Holywood3d 		& $1002$	& $717$		& $469$		& $447$		& $330980$		& $28$	& $\mathbf{0.48}$	& $0.57$ \\ \hline
				Olympic Sports 	& $699$		& $1063$	& $751$		& $220$		& $257926$		& $29$	& $\mathbf{0.18}$	& $0.38$ \\ \hline
				CMC 			& $274$		& $300$		& $477$		& $114$		& $334599$		& $21$	& $0.25$	& $\mathbf{0.18}$ \\ \hline
				PIMA 			& $217$		& $551$		& $364$		& $441$		& $183362$		& $14$	& $0.85$	& $\mathbf{0.12}$ \\ \hline
				YEAST 			& $312$		& $323$		& $461$		& $484$		& $395465$		& $33$	& $3.05$	& $\mathbf{0.33}$ \\ \hline
				
			\end{tabularx}
		}
	\end{center}
\end{table*}

\begin{table*}[t!]
	\begin{center}
		\caption{Computational complexity during inference (FLOPS)}\label{t5}
		\resizebox{\textwidth}{!}{
			\begin{tabularx}{\textwidth}{ |c| *{8}{Y|} }\cline{2-9}
				\multicolumn{1}{c|}{}
				
				& HeMLGOP 		& HoMLGOP		& HeMLRN			& HoMLRN 		& POP		& PMLP 		& PLN		& BLS	\\ \hline
				Holywood3d 		& $31.4$k	& $25.1$k	& $42.2$k	& $66.8$k	& $113.3$k	& $46.4$k	& $\mathbf{13.5}$k	& $157.1$k \\ \hline
				Olympic Sports 	& $31.1$k	& $22.1$k	& $62.1$k	& $52.2$k	& $206.7$k	& $208.8$k	& $\mathbf{14.6}$k	& $118.2$k \\ \hline
				CMC 			& $2.5$k	& $\mathbf{1.6}$k	& $9.0$k	& $2.6$k	& $248.2$k	& $167.2$k	& $10.0$k	& $14.4$k	 \\ \hline
				PIMA  			& $\mathbf{2.2}$k	& $4.8$k	& $4.6$k	& $54.8$k	& $126.4$k	& $85.6$k	& $43.5$k	& $13.4$k \\ \hline
				YEAST  			& $\mathbf{1.9}$k 	& $2.8$k	& $6.4$k	& $18.2$k	& $55.2$k	& $88.8$k	& $728.5$k	& $382.3$k \\ \hline \hline
				15 scenes  		& $\mathbf{248.0}$k	& $356.8$k	& $290.5$k	& $413.0$k	& -			& $373.2$k	& $355.1$k	& $853.7$k  \\ \hline
				MIT indoor  	& $\mathbf{152.0}$k 	& $225.0$k	& $372.6$k	& $321.4$k	& - 		& $232.5$k	& $289.8$k	& $1563.1$k \\ \hline
				Caltech101  	& $187.0$k 	& $344.6$k	& $338.8$k	& $\mathbf{124.3}$k	& - 		& $408.2$k	& $417.3$k	& $1883.1$k \\ \hline
				Caltech256  	& $\mathbf{175.3}$k 	& $461.6$k	& $380.4$k	& $462.1$k	& - 		& $470.5$k	& $4083.4$k	& $5323.3$k \\ \hline
				PubFig 			& $150.5$k 	& $\mathbf{130.0}$k	& $294.1$k	& $324.6$k	& - 		& $319.8$k	& $316.9$k	& $1698.1$k \\ \hline
				CFW60k 			& $250.2$k 	& $\mathbf{184.6}$k	& $204.9$k	& $205.1$k	& - 		& $406.6$k	& $15449.4$k	& $11182.9$k \\ \hline
				
			\end{tabularx}
		}
	\end{center}
\end{table*}

Regarding the hyper-parameter settings of the proposed method (HeMLGOP) and other variants (HoMLRN, HeMLRN, HoMLGOP), $c \in \{10^{-1}, 1, 10^{1}\}$ was used in the Randomized Network step. For all the methods that employ BP, it is important to properly regularize the network structure to avoid overfitting. Regularization setting in BP includes weight regularization and Dropout \cite{srivastava2014dropout}. We experimented with $2$ types of weight regularization: weight decay with scale of $0.0001$ and $l_2$ norm constraint with maximum value in $\{1.0, 2.0, 3.0\}$. The dropout step was applied to the output of the hidden layer with the percentage selected from $\{0.5,0.3,0.1\}$. In addition, $0.2$ dropout was applied to the deep feature input during BP. For small-scale problems, during progressive learning, the following learning rate schedule $\{0.01, 0.001,0.0001\}$ and the corresponding number of epochs $\{20,40,40\}$ were applied to all methods while in medium/large scale datasets, the learning rate schedule and the number of epochs were set to $\{0.01, 0.005, 0.001,0.0005, 0.0001, 0.00001\}$, $\{100,100,100,100,100\}$, respectively. During fine-tuning stage, all networks were fine-tuned for $200$ epochs with the fixed learning rate $0.00005$.

For POP and PMLP, we defined a network template of maximum $4$ layers and $8$ layers in small-scale and medium/large scale problems, respectively, with each layer having $200$ neurons. In all competing algorithms that have incremental steps within a layer, the layer starts with $40$ neurons and increments $20$ neurons at each step, i.e., $n_{min} = 40, n_i = 20$. To avoid the problem of growing arbitrarily large hidden layers, and to make the learned architectures comparable between all methods, we limit the maximum number of added neurons in a hidden layer in our proposed method and $3$ other variants to $200$, and $300$ for PLN and BLS. Moreover, we applied a universal convergence criterion based on the rate of improvement during network progression for all methods, i.e., an algorithm stops adding neurons to the current hidden layer when $r_i < 10^{-4}$ and stops adding hidden layers when $r_l < 10^{-4}$ with $r_i$ and  $r_l$ calculated according to Eq. (\ref{eq7}) and (\ref{eq8}) with accuracy as the performance measure.

\subsection{Results}

Table \ref{t3} shows the classification accuracy of all competing methods on the $11$ datasets and Table \ref{t6} shows the corresponding model sizes, i.e., the number of parameters in the network.

Regarding the performances on small-scale datasets, among all competing algorithms, it is clear that the proposed algorithm (HeMLGOP) and its heterogeneous variant (HeMLRN) are the leading performers. The differences, in terms of classification accuracy, between the two algorithms are statistically insignificant. However, the models learned by HeMLGOP are significantly smaller (approximately $3\times$ in most cases) as compared to those learned by HeMLRN. Between two homogeneous variants of the proposed algorithm, it is clear that HoMLGOP consistently outperforms HoMLRN in terms of both classification accuracy and compactness.

Between the two algorithms that employ BP during progressive learning, the results of HeMLGOP are similar or better than its homogeneous counterpart (HoMLGOP) in both classification accuracy and memory footprint, with the only exception in Holywood3d in which HeMLGOP is $2\%$ more accurate, requiring $2.4k$ more parameters. The empirical results of HeMLGOP and its $3$ variants are consistent with our discussion in Section \ref{proposed_discussion}: allowing heterogeneous neurons within a hidden layer can lead to better performing networks, and weights adjustment through BP is necessary to fully harness the representation power of newly added neurons during progressive learning.

\begin{figure*}[t!]
	\centering
	\includegraphics[width=0.95\textwidth]{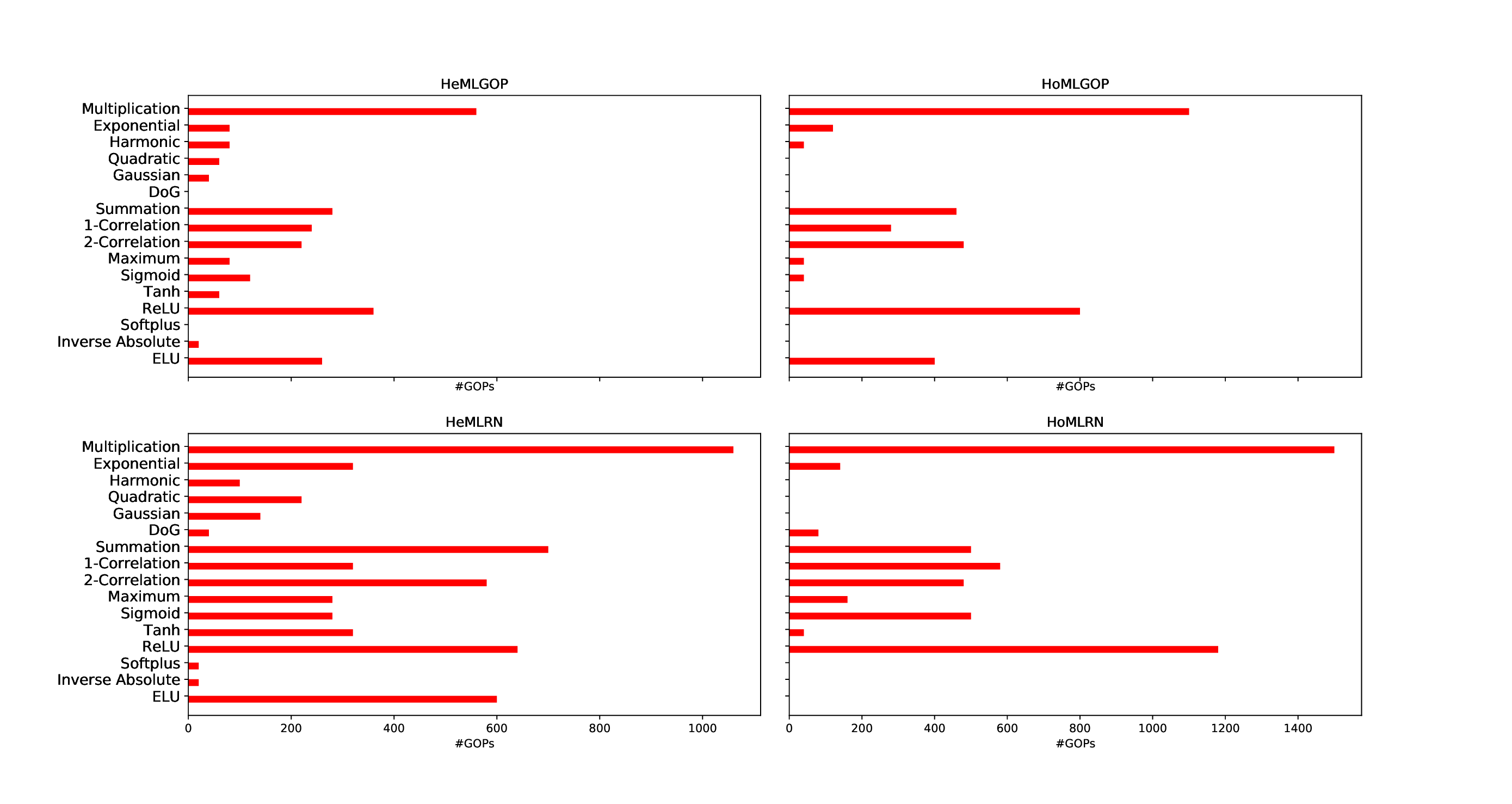}
	\caption{Operator distribution}
	\label{f2}
\end{figure*}

Regarding the performance of POP and PMLP on small-scale datasets, it is obvious that the final network topologies learned by the two algorithms are enormous, compared to the proposed algorithm and its variants. Particularly, in CMC, PIMA, and YEAST, HeMLGOP needs approximately only $1000$ parameters, while POP and PMLP require a vast amount of more than $40k$ parameters, i.e., $40\times$ memory saving achieved by HeMLGOP with similar or better accuracies. The differences between POP, PMLP and the proposed algorithm are expected since our proposed algorithm addresses the two limitations in POP as discussed in Section \ref{pop_limitations}: fixed hidden layer sizes and the homogeneity constraint of a layer. Regarding PLN, while the algorithm requires slightly fewer parameters in Holywood3d and Olympic Sports datasets as compared with HeMLGOP, the classification performances of PLN are similar or much worse. In other small-scale problems, PLN is inferior to HeMLGOP in both accuracy and compactness.

Similar phenomena between the competing algorithms can be observed in medium/large scale datasets: the proposed HeMLGOP remains as the best performing algorithm to learn the most compact network topologies while being similarly or more accurate than other benchmarked algorithms. The classification accuracies of HeMLRN and HoMLGOP, are competitive with the proposed algorithm, however, achieved by larger network configurations. HoMLRN performs worst among its GOP-based counterparts. While being as accurate as the proposed HeMLGOP in most medium/large scale problems, the models learned by PMLP require $2\times$ to $3\times$ number of parameters. Moving to a medium/large scale setting with more challenging problems, PLN and BLS are consistently inferior to other algorithms in both measures. In addition, the networks grown by PLN in some datasets such as Caltech256 ($257$ classes) or CFW ($500$ classes) are enormous with the number of parameters reaching the order of millions. This is due to the limitations in PLN that the size of a hidden layer is always equal or larger than twice the number of target outputs since a new hidden layer is constructed based on the current output layer. By having only $2$ hidden layers and resorting entirely to random weights during progression, BLS tends to grow large but inefficient networks as compared to other algorithms.

As mentioned in Section \ref{proposed_method}, one of the motivations in our work is to speed up the operator set searching procedure in GOP-based system like POP. Thus, Table \ref{t4} presents the training time (in seconds) of all algorithms on small-scale problems. It should be noted that algorithms based on BP can take huge advantage of GPUs to speed up the training process. However, to give comparable results in terms of training time of all competing methods, we conducted all small-scale experiments based on CPUs with the same machine configuration. It is clear that the proposed algorithm is much faster than POP by approximately $300\times$ in most cases. While HoMLGOP, HeMLRN, HoMLRN can be seen as simplified versions of the proposed algorithm, there is no clear winner among the four algorithms in the context of training time. Depending on the difficulty of the given problem, the training time of HeMLGOP is relatively short as compared to its variants since the proposed algorithm tends to converge after only a few progressive steps with small network topologies, e.g., in CMC, PIMA, YEAST. Among all competing algorithms, it is clear that PLN and BLS are the fastest algorithms to train since both algorithms rely only on random weights during the network progression. As shown in Table \ref{t3}, this advantage of fast training time results in the cost of inferior performances and very large model sizes for deployment as compared to the proposed algorithm. While Table \ref{t4} can give an intuition on the speed of each algorithm during the training stage, it is worth noting that the benchmark can only give strict comparisons between the proposed algorithm, its variants, and POP, all of which are based on our unoptimized implementation of GOP. The exact relative comparison on the training time between perceptron-based networks (PMLP, PLN, BLS) and GOP-based networks (POP, the proposed algorithm and, its variants) can change drastically when an optimized implementation of GOP is available.

Nowadays, with the development of commodity hardware, training an algorithm in some orders of magnitude longer than another might not prevent its application. However, deploying a large pretrained model to constrained computing environments such as those in mobile, embedded devices poses a big challenge in practical application. Not only the storage requirement plays an important factor during the deployment stage in mobile devices but also the amount of energy consumed during inference. While the actual inference time of each algorithm depends on the details of the implementation such as hardware-specific optimizations or concurrency support, the computational complexity of an algorithm is directly related to the energy consumption. Under this perspective, Table \ref{t5} shows the number of floating-point operations (FLOPs) required by each network in Table \ref{t3} to predict a new sample. With compact network configurations, it is clear that $5$ out of $11$ datasets, the proposed algorithm requires the smallest number of operations during inference. In other cases, the number of FLOPs in HeMLGOP remains relatively low as compared to other algorithms. For example, in CMC, Caltech101 or PubFig, HeMLGOP is the second best in terms of computational complexity. Although being very fast to train, making inference with the learned models produced by PLN and BLS is costly in most cases. For example, in YEAST, Caltech256 and CFW, the numbers of FLOPs in PLN and BLS are more than $200\times$ compared to HeMLGOP. Regarding POP, not only does the algorithm take a very long time to train but also heavy costs to make an inference. The computational complexity of PMLP during testing is on the same order as POP, however, with much shorter training time. It is worth noting that in case of GOP-based networks, the number of parameters in the model and the inference cost are not directly related, i.e., two networks having the same topology could have different inference complexities. This is due to the fact that different operator sets in GOP possess different computational complexities.

Figure \ref{f2} shows the distribution of operators used by our proposed algorithm and its three variants in all datasets. It is clear that while the proposed algorithm and its heterogeneous variant (HeMLRN) used a diverse set of operators, the types of operators selected by HoMLGOP and HoMLRN are more limited. Within the library of nodal operators, ``Multiplication" was popular among all four algorithms. Similar observations can be seen in activation operators: ``ReLU" and ``ELU" were favored as activation functions while ``Summation", ``1-Correlation" and ``2-Correlation" were the most popular pooling operators to all algorithms.

\section{Conclusions}
In this paper, we proposed an efficient algorithm to learn fully heterogeneous multilayer networks that utilize Generalized Operational Perceptron. The proposed algorithm (HeMLGOP) is capable of learning very compact network structures with efficient performances. Together with the proposed algorithm, we also presented $3$ related variants which can be seen as simplified versions of HeMLGOP. Extensive experimental benchmarks in real-world classification problems have shown that, under different perspectives, the proposed algorithm and its variants outperform other related approaches, including POP, an algorithm that also employs Generalized Operational Perceptron.

\appendices

\section{Computational Complexity}\label{A1}
Since the proposed algorithm determines the network's topology both in block-wise and layer-wise manner, after training a hidden layer, the training data can always be transformed by calculating the outputs of that layer to produce the new training data to be used for determining the next hidden layer. Therefore, when estimating the computational complexity at each progressive step, we consider the complexity under a Single Hidden Layer Network (SHLN) configuration with the specifications given in Table \ref{t7}.

\begin{table}[t]
	\begin{center}
		\caption{SHLN specification}\label{t7}
		\resizebox{0.7\linewidth}{!}{
			\begin{tabular}{|c|c|}\hline
				
				Parameter 			& Notation			\\ \hline \hline
				Input dimension		& $D_I$ \\ \hline
				Output dimension	& $D_O$ \\ \hline
				\# current hidden neurons & $D_H$ \\ \hline
				\# new hidden neurons 	 & $D_h$ \\ \hline
				\# training samples		 & $N$ \\ \hline
				\# operator sets			 & $N_{\mathbf{O}}$ \\ \hline	
				\# BP epochs				 & $E$ \\ \hline
				
			\end{tabular}
		}
	\end{center}
\end{table}

Since the computational complexity of each GOP neuron depends on the form of the operator set and the input dimensions, we denote the complexity of the $i$-th GOP neuron as a function of number of input dimensions: $\mathcal{F}_{i}(D_I)$ and $\mathcal{B}_{i}(D_I)$ for the forward and backward pass respectively. At each progressive step, the algorithm consists of two procedures: the search of the operator set for $D_h$ neurons through $N_{\mathbf{O}}$ randomized processes and the weights and biases update through BP.

\begin{itemize}
\item Complexity of a randomized process: this involves producing the hidden representation of the training data $O\big(N\sum_{i=1}^{D_H + D_h} \mathcal{F}_i(D_I)\big)$ and solving the least-square problem $O\big((D_H + D_h)^3\big)$ (here we assume that $D_H + D_h < N$ as usually the case).

\item Complexity of weights and biases update through BP: in an epoch, the forward pass involves producing the hidden representation $O\big(N\sum_{i=1}^{D_H + D_h} \mathcal{F}_i(D_I)\big)$ and the output representation $O\big(N(D_H + D_h)D_O\big)$, the backward pass involves updating the weights and biases of $D_h$ GOPs $O\big(N\sum_{i=D_{H}+1}^{D_{H} + D_{h}} \mathcal{B}_i(D_I)\big)$ and the linear output layer $O\big(2N(D_H + D_h)D_O + (D_H + D_h)D_O \big)$ (here we assume the MSE loss function).

\end{itemize}

To conclude, given $N_{\mathbf{O}}$ operator sets and $E$ BP epochs, the estimated computational complexity of the given progressive step is $O\bigg(N_{\mathbf{O}}\big( N\sum_{i=1}^{D_H + D_h} \mathcal{F}_i(D_I) + (D_H + D_h)^3\big) + E\big(N\sum_{i=1}^{D_H + D_h} \mathcal{F}_i(D_I) + 3N(D_H + D_h)D_O + N\sum_{i=D_{H}+1}^{D_{H} + D_{h}} \mathcal{B}_i(D_I) + (D_H + D_h)D_O \big)\bigg)$

\section{Proof of Convergence}\label{A2}

To prove that progressive learning in each layer of HeMLGOP converges, we adopt the following notations:

\begin{itemize}

\item $\mathbf{h}^{(l)}_{k} \in \mathbb{R}^{N \times D^{(l)}_k}$ denote the output produced by the $k$-th block of the $l$-th hidden layer given the training data.
\item $\mathbf{H}^{(l)}_{k} = [\mathbf{h}^{(l)}_{1}, \dots, \mathbf{h}^{(l)}_{k}] \in \mathbb{R}^{N \times (D^{(l)}_1 + \dots + D^{(l)}_k)}$ denote the output produced by the first $k$ blocks of the $l$-th hidden layer, given the training data.
\item $\mathbf{B}^{(l)}_{k} \in \mathbb{R}^{(D^{(l)}_1 + \dots + D^{(l)}_k) \times C}$ denotes the output weights obtained when adding the $k$-th block in the $l$-th hidden layer.
\item $E^{(l)}_{k} \in \mathbb{R}$ denotes the training Mean Squared Error (MSE) obtained with $\mathbf{H}^{(l)}_{k}$ and $\mathbf{B}^{(l)}_{k}$. 

\end{itemize}

We will prove that the sequence $(E^{(l)}_{k})_{\{k\}}$ is monotonically decreasing. By using the fact that $(E^{(l)}_{k})_{\{k\}}$ is bounded below by $0$, we can then conclude that $(E^{(l)}_{k})_{\{k\}}$ converges.

Firstly, we should note that $\mathbf{B}^{(l)}_{k}$ is obtained after the intermediate fine-tuning step when adding the $k$-th block. Given a fixed $\mathbf{H}^{(l)}_{k}$, $\mathbf{B}^{(l)}_{k}$ is not necessarily the optimal weights in terms of MSE since we have the following relation:

\begin{equation}\label{eq9}
E^{(l)}_{k} = \|\mathbf{H}^{(l)}_{k}\mathbf{B}^{(l)}_{k} - \mathbf{Y}\|_2^2 \geq \|\mathbf{H}^{(l)}_{k}\mathbf{B^{*}}^{(l)}_{k} - \mathbf{Y}\|_2^2
\end{equation}
where $\mathbf{B^{*}}^{(l)}_{k} = (\mathbf{H}^{(l)}_{k})^{\dagger}\mathbf{Y}$ denote the optimal weights obtained by the least-square solution.

When adding the $(k+1)$-th block in the $l$-th hidden layer, after the randomized operator set search, we obtain the hidden representation $\mathbf{H}^{(l)}_{k+1} = [\mathbf{H}^{(l)}_{k}, \mathbf{h}^{(l)}_{k+1}]$ in which $\mathbf{H}^{(l)}_{k}$ is fixed from the previous progressive step and $\mathbf{h}^{(l)}_{k+1}$ is generated by the new random GOPs, and the corresponding optimal output weights $\mathbf{b}^{(l)}_{k+1} \in \mathbb{R}^{(D^{(l)}_1 + \dots + D^{(l)}_{k+1}) \times C}$ with respect to $\mathbf{H}^{(l)}_{k+1}$ in terms of MSE. Thus

\begin{align}
\mathcal{E}^{(l)}_{k+1} &= \|\mathbf{H}^{(l)}_{k+1}\mathbf{b}^{(l)}_{k+1} - \mathbf{Y}\|_2^2 \leq \|\mathbf{H}^{(l)}_{k+1}\mathbf{b} - \mathbf{Y}\|_2^2 \label{eq10} \\
& \forall  \mathbf{b} \in \mathbb{R}^{(D^{(l)}_1 + \dots + D^{(l)}_{k+1}) \times C}
\end{align}
where $\mathcal{E}^{(l)}_{k+1}$ denote the training MSE after the operator set searching step when adding the $(k+1)$-th block.

Since (\ref{eq10}) holds for all $\mathbf{b}$, we can replace $\mathbf{H}^{(l)}_{k+1}$ with $[\mathbf{H}^{(l)}_{k}, \mathbf{h}^{(l)}_{k+1}]$ and $\mathbf{b}$ with \myvec{\mathbf{B^{*}}^{(l)}_{k}\\ \mathbf{0}} to have the following relation

\begin{equation}
\mathcal{E}^{(l)}_{k+1} \leq \|[\mathbf{H}^{(l)}_{k}, \mathbf{h}^{(l)}_{k+1}] \myvec{\mathbf{B^{*}}^{(l)}_{k}\\ \mathbf{0}} - \mathbf{Y} \|_2^2
\end{equation}
or
\begin{align}
\mathcal{E}^{(l)}_{k+1} & \leq \|\mathbf{H}^{(l)}_{k}\mathbf{B^{*}}^{(l)}_{k} + \mathbf{h}^{(l)}_{k+1}\mathbf{0} - \mathbf{Y} \|_2^2 \\
& = \|\mathbf{H}^{(l)}_{k}\mathbf{B^{*}}^{(l)}_{k} - \mathbf{Y} \|_2^2 \leq E^{(l)}_{k} \label{eq11}
\end{align}

Since $E^{(l)}_{k+1}$, the MSE achieved after adding the $(k+1)$-th block, is obtained by fine-tuning $\mathbf{h}^{(l)}_{k+1}$ and $\mathbf{b}^{(l)}_{k+1}$ (which becomes $\mathbf{B}^{(l)}_{k}$) through stochastic gradient descent which has been proven to converge to the local optimum \cite{robbins1985stochastic} with small enough step sizes, we have the following relation:

\begin{equation}\label{eq12}
E^{(l)}_{k+1} \leq \mathcal{E}^{(l)}_{k+1}
\end{equation}

From (\ref{eq11}) and (\ref{eq12}), we have $E^{(l)}_{k+1} \leq E^{(l)}_{k}$ thus the sequence $(E^{(l)}_{k})_{\{k\}}$ is monotonically decreasing. \QEDA

We should note that by discarding the unrelated steps, the above proof is also valid for the three variants of HeMLGOP.

\section{Training Curves of The Proposed Algorithm}\label{A3}

\begin{figure}[h]
	\centering
	\includegraphics[width=1.0\linewidth]{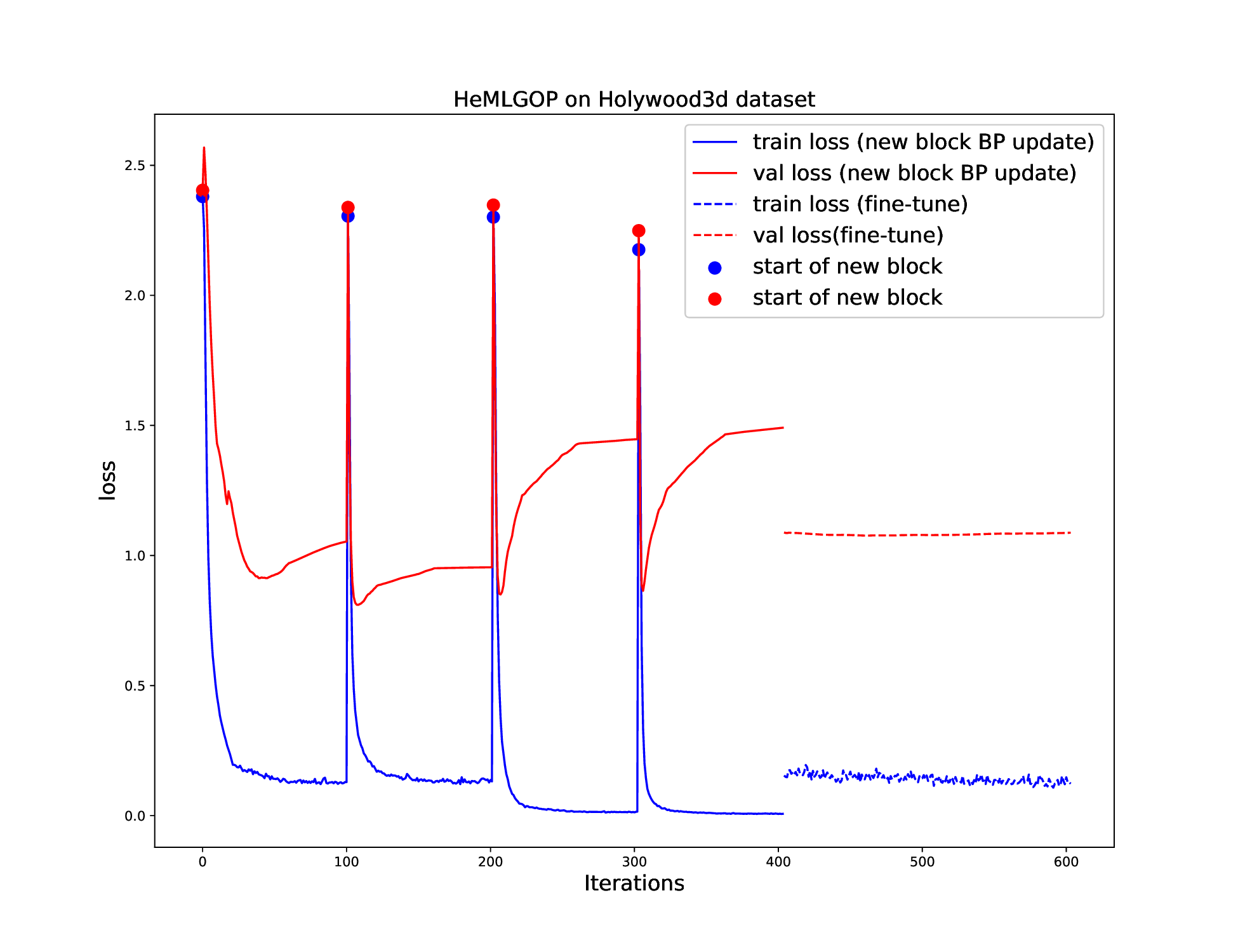}
	\caption{Loss curve of HeMLGOP on Holywood3d}
	\label{f3}
\end{figure}

\begin{figure}[h]
	\centering
	\includegraphics[width=1.0\linewidth]{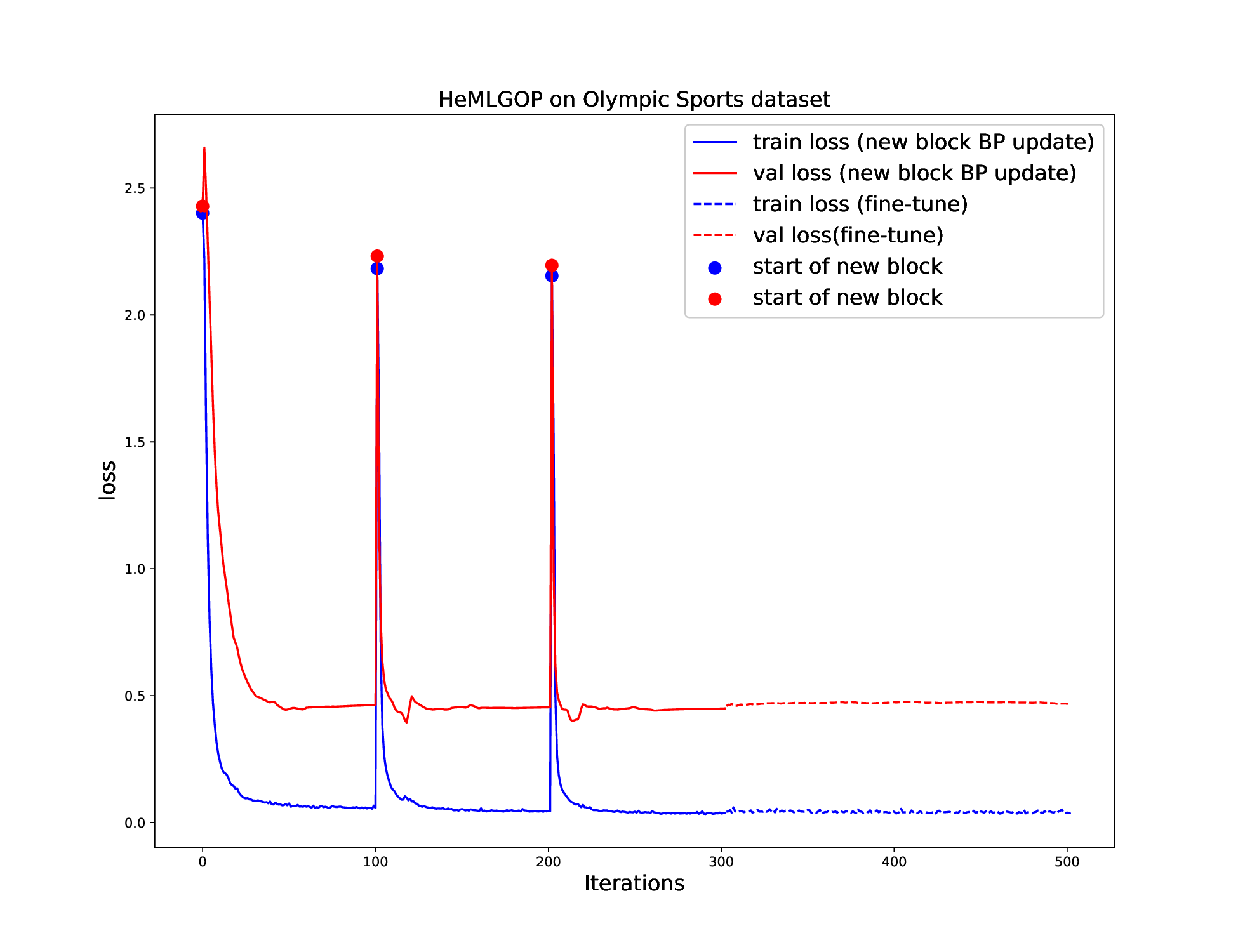}
	\caption{Loss curve of HeMLGOP on Olympic Sports}
	\label{f4}
\end{figure}

\begin{figure}[h]
	\centering
	\includegraphics[width=1.0\linewidth]{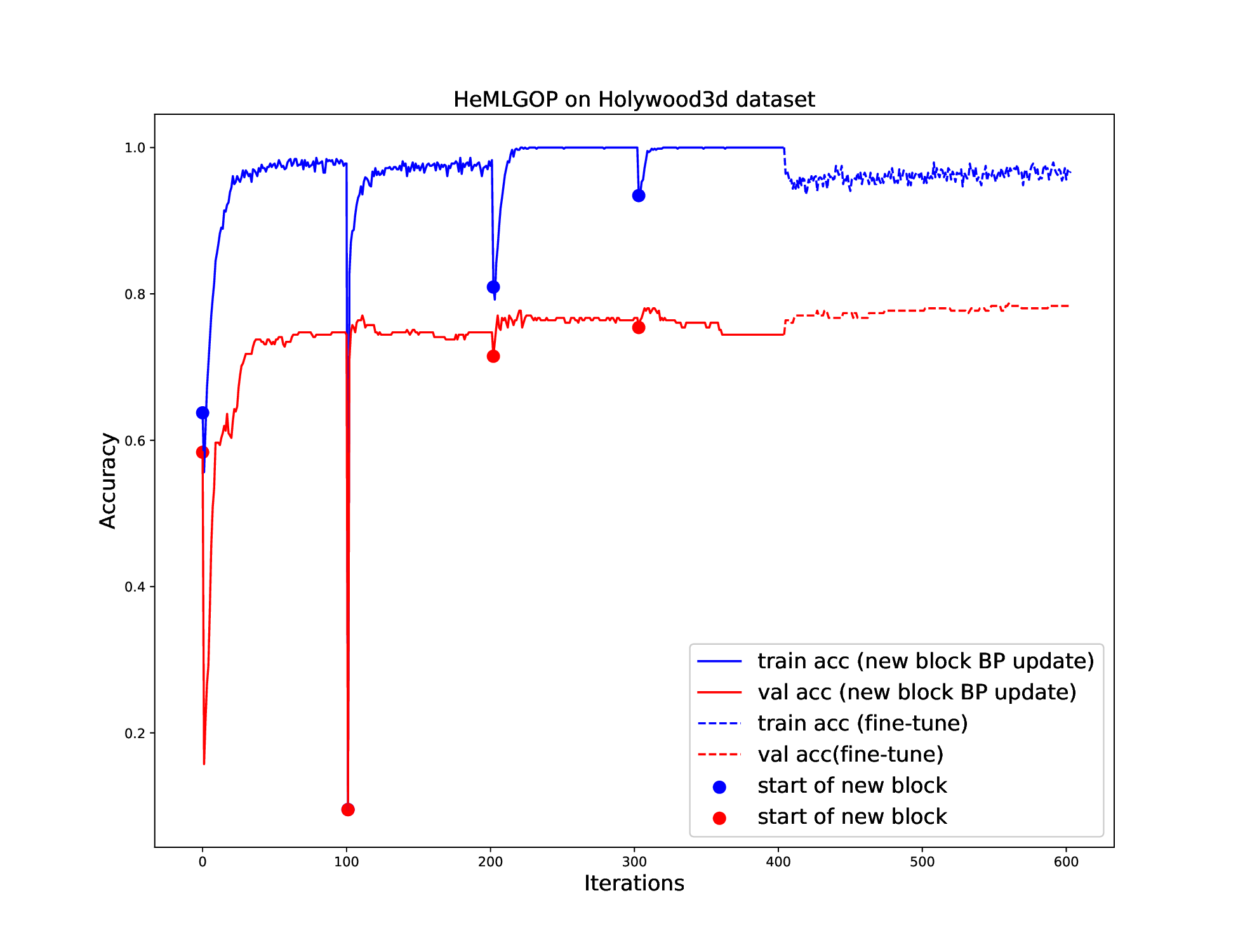}
	\caption{Accuracy curve of HeMLGOP on Holywood3d}
	\label{f5}
\end{figure}

\begin{figure}[h]
	\centering
	\includegraphics[width=1.0\linewidth]{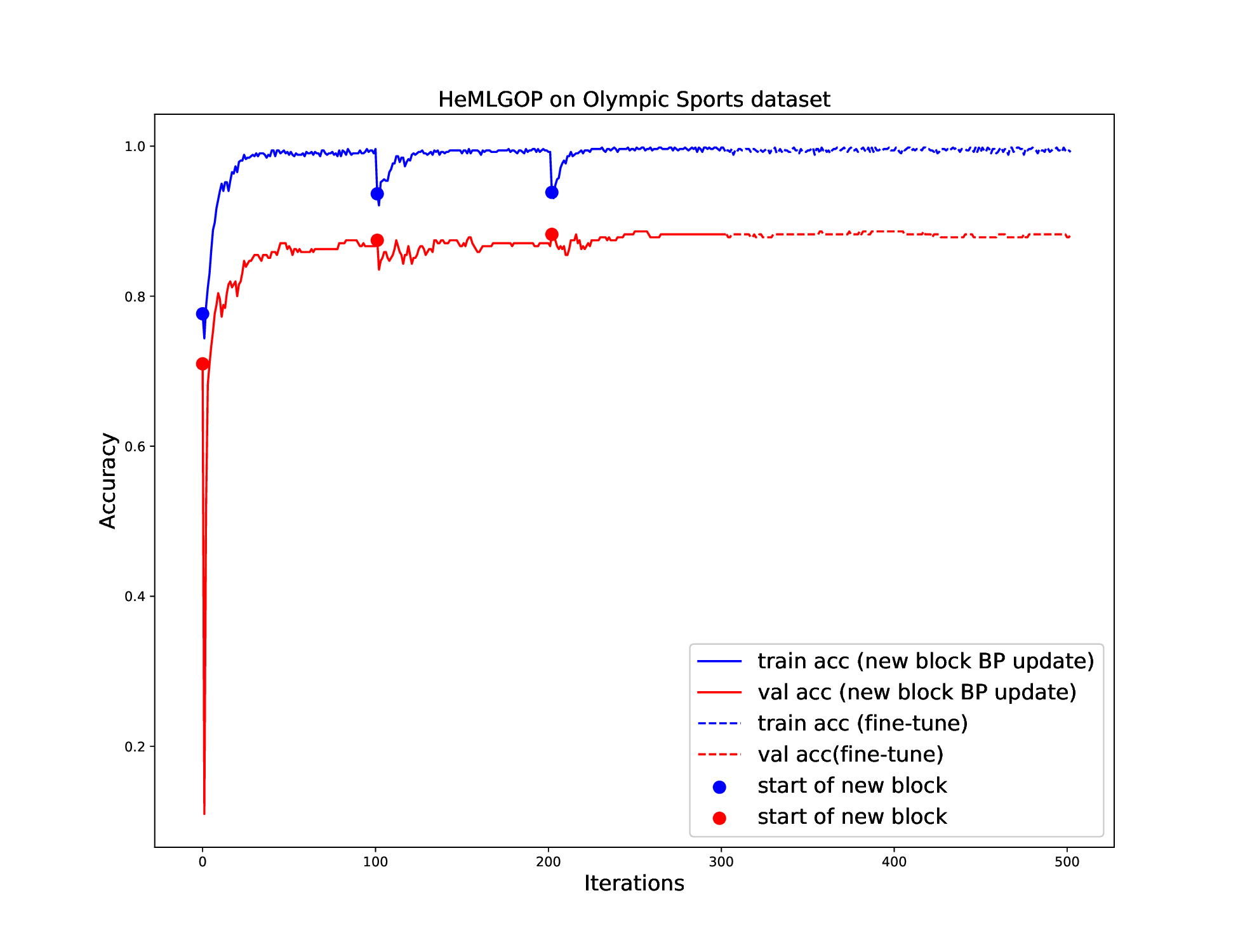}
	\caption{Accuracy curve of HeMLGOP on Olympic Sports}
	\label{f6}
\end{figure}

The training curves of HeMLGOP on Holywood3D and Olympic Sports are shown in Figures \ref{f3}-\ref{f6}. In all figures, the continuous lines show the statistics (accuracy/loss) during Back Propagation at progressive learning stage while the dashed lines show the statistics during fine-tuning stage of the whole network. Here it should be noted that we did not attempt to tune the hyper-parameters during the fine-tuning stage but we used a small (and fixed) learning rate $0.00005$ to fine-tune all the networks for $200$ epochs. In practical cases, to better take advantage of the fine-tuning stage, one could further tune hyper-parameters such as learning rate and learning rate schedule following a trial and error approach (as is the standard approach) for individual datasets. However, since this is not practical for a systematic comparison of many competing methods, as is the case of the study in our experiments, we relied on a basic approach using the fixed learning rate.

\bibliography{reference}
\bibliographystyle{ieeetr}

\end{document}